%% file: aaai25.tex
\documentclass[letterpaper]{article} 
\usepackage{aaai25}

\usepackage{times}  
\usepackage{helvet}  
\usepackage{courier}  
\usepackage[hyphens]{url}  
\usepackage{graphicx} 
\usepackage{natbib}  

\usepackage{caption} 
\usepackage{algorithm}
\usepackage{algorithmic}
\usepackage{tikz}
\graphicspath{{figures/}}
\usepackage{rotating}
\usepackage{amsmath}
\usetikzlibrary{arrows.meta}
\usepackage{amsthm}
\usepackage{times}
\usepackage{helvet}
\usepackage{courier}
\usepackage{graphicx}
\usepackage{amsmath}
\usepackage{amssymb}
\usepackage{booktabs}
\usepackage{multirow}
\usepackage{makecell}
\usepackage{subfigure}
\usepackage{subcaption}
\usepackage{placeins}
\usepackage{caption}

\newtheorem{definition}{Definition}
\newtheorem{proposition}{Proposition}

\usepackage{newfloat}
\usepackage{listings}
\DeclareCaptionStyle{ruled}{labelfont=normalfont,labelsep=colon,strut=off} 
\floatstyle{ruled}
\newfloat{listing}{tb}{lst}{}
\floatname{listing}{Listing}
\nocopyright

\title{ScaleNet: Scale Invariance Learning in Directed Graphs}
\author {
    Qin Jiang,
    Chengjia Wang,
    Michael Lones,
    Yingfang Yuan,
    Wei Pang
}
\affiliations {
    School of Mathematical and Computer Sciences, Heriot Watt University\\
 W.Pang@hw.ac.uk
}

\usepackage{bibentry}

\usetikzlibrary{arrows.meta, positioning}
\usepackage{fancyhdr} 
\usepackage{amsmath}
\usepackage{amsthm}

\begin{document}

\maketitle
\input{0_0_abstract}
\input{0_1_introduction}

\input{0_2_motivation_considerations}

\input{1_1_preliminary}

\input{1_2_methodology}

\input{conclusions}  
\bibliography{aaai25}

\clearpage
\newpage
\appendix
\setcounter{table}{5}
\setcounter{figure}{3}

\input{appendix_data}

\input{appendix_technical}

\input{appendix_technical_inception}

\input{reproducibility_checklist}
\end{document}

%% file: 0_0_abstract.tex
\begin{abstract}

Graph Neural Networks (GNNs) have advanced relational data analysis but lack invariance learning techniques common in image classification. In node classification with GNNs, it is actually the ego-graph of the center node that is classified. This research extends the scale invariance concept to node classification by drawing an analogy to image processing: just as scale invariance being used in image classification to capture multi-scale features, we propose the concept of ``scaled ego-graphs''. Scaled ego-graphs generalize traditional ego-graphs by replacing undirected single-edges with ``scaled-edges'', which are ordered sequences of multiple directed edges. We empirically assess the performance of the proposed scale invariance in graphs on seven benchmark datasets, across both homophilic and heterophilic structures. Our scale-invariance-based graph learning outperforms inception models derived from random walks by being simpler, faster, and more accurate. The scale invariance explains inception models' success on homophilic graphs and limitations on heterophilic graphs. To ensure applicability of inception model to heterophilic graphs as well, we further present ScaleNet, an architecture that leverages multi-scaled features. ScaleNet achieves state-of-the-art results on five out of seven datasets (four homophilic and one heterophilic) and matches top performance on the remaining two, demonstrating its excellent applicability. This represents a significant advance in graph learning, offering a unified framework that enhances node classification across various graph types. Our code is available at 
https://github.com/Qin87/ScaleNet/tree/July25.


\end{abstract}

%% file: 0_1_introduction.tex
\section{Introduction}

Graph Neural Networks (GNNs) have become prominent for learning from graphs. Recently, there has been a surge of interest in developing GNNs specifically tailored for directed graphs (digraphs). Digraphs are prevalent in real-world scenarios where nodes are connected in a one-directional manner, providing more valuable directional information than undirected graphs.

GNNs are generally classified into spatial and spectral methods. Extending spatial GNNs to digraphs is relatively straightforward for propagating information over directed edges. However, this leads to information flow in only one direction, potentially overlooking valuable data from the opposite direction \cite{rossiEdgeDirectionalityImproves2023}. Additionally, the asymmetric graph Laplacian of a digraph prevents orthogonal eigendecomposition \cite{zhangMagNetNeuralNetwork}, rendering traditional spectral methods \cite{defferrardConvolutionalNeuralNetworks2016} ineffective. These challenges underscore the need for innovative approaches to better exploit the rich structural information inherent in directed graphs.

There are three main classes of GNN extensions designed for digraphs:

\begin{enumerate}
    \item \textbf{Real Symmetric Laplacians}: 
   These methods extract symmetric representations of directed graphs to effectively capture directional features. For instance, MotifNet \cite{montiMOTIFNETMOTIFBASEDGRAPH2018} constructs a symmetric adjacency matrix based on motifs, though this approach is constrained by the need for predefined templates and struggles with complex structures. DGCN \cite{maSpectralbasedGraphConvolutional2019}, SymDiGCN \cite{tongDirectedGraphConvolutional2020}, and DiGCN \cite{tongDigraphInceptionConvolutional2020} address this by incorporating both the asymmetric adjacency matrix and its transpose from a Markov process perspective. However, these methods are computationally intensive \cite{kolliasDirectedGraphAutoEncoders2022}, limiting their scalability to large graphs.

    \item \textbf{Hermitian Laplacians}: 
 These methods utilize complex-valued entries in Hermitian matrices to encode directional information while retaining positive semi-definite eigenvalues. MagNet \cite{zhangMagNetNeuralNetwork} uses Hermitian matrices with complex numbers for this purpose. Extensions to signed digraphs include SigMaNet \cite{fioriniSigMaNetOneLaplacian2023} and MSGNN \cite{heMSGNNSpectralGraph2022}. QuaNet \cite{fioriniGraphLearning4D2023} introduced Quaternion-valued Laplacians to capture asymmetric weights in signed directed graphs.
  

 \item \textbf{Bidirectional Spatial Methods}: 
 Recent advancements include Dir-GNN \cite{rossiEdgeDirectionalityImproves2023}, which learns from in-neighbors and out-neighbors separately to capture directional information. Similar approaches are explored in \cite{zhuoCommuteGraphNeural2024}.
 This approach demonstrates notable improvements on heterophilic graphs but is somewhat less effective on homophilic graphs. 
\end{enumerate}

While Real Symmetric Laplacian and Hermitian Laplacian methods are generally effective for homophilic graphs, they often underperform on heterophilic graphs. Moreover, Real Symmetric Laplacian methods are computationally expensive, making them less scalable. On the other hand, Bidirectional Spatial Methods excel in heterophilic graph settings but are less optimal for homophilic graphs.

In this research, we introduce a unified network that performs well on both homophilic and heterophilic graphs. By transforming the original graph into multiple scaled versions and demonstrating scale invariance, we synthesize these graphs based on dataset-specific characteristics. Optional components, such as self-loops, batch normalization, and non-linear activation functions, are also incorporated to enhance performance. Our method achieves state-of-the-art results on five out of seven datasets (four homophilic and one heterophilic) and matches top performance on the remaining two, demonstrating its robust applicability and effectiveness.

Compared to real symmetric Laplacian methods, our approach is simpler, faster, and delivers superior performance. Compared to single-scale methods like MagNet \cite{zhangMagNetNeuralNetwork} and Dir-GNN \cite{rossiEdgeDirectionalityImproves2023}, our method proves to be more robust, particularly in the context of imbalanced data.

%% file: 0_2_motivation_considerations.tex
\section{Motivation and Related Work}

\subsection{Invariant Learning Techniques}

Invariant classifiers generally exhibit smaller generalization errors compared to non-invariant techniques \cite{wuHandlingDistributionShifts2022}. Therefore, explicitly enforcing invariance in GNNs could potentially improve their robustness and accuracy \cite{sokolicGeneralizationErrorInvariant2017b}. To improve the generalization ability of GNNs, in this research we focus on invariant learning techniques for node classification.

An invariant classifier \cite{sokolicGeneralizationErrorInvariant2017b} is less affected by specific transformations of the input than non-invariant classifiers. Invariant learning techniques have been well studied in Convolutional Neural Networks (CNNs), where they address translation, scale, and rotation invariance for image classification \cite{cohen2016group, Lenc_2015_CVPR}.

However, the application of invariant classifiers in GNNs is less explored. The non-Euclidean nature of graphs introduces extra complexities that make it challenging to directly achieve invariance, and thus the invariance methods derived from CNNs cannot be straightforwardly applied.

\subsection{Invariance of Graphs}

In Graph Neural Networks (GNNs), two types of permutation invariance are commonly utilized:

\begin{enumerate}
\item \textbf{Global Permutation Invariance}:
Across the entire graph, the output of a GNN should remain consistent regardless of the ordering of nodes in the input graph. This property is particularly useful for graph augmentation techniques \cite{xieArchitectureAugmentationPerformance2024}.
\item \textbf{Local Permutation Invariance}:
At each node, permutation-invariant aggregation functions ensure that the results of operations remain unaffected by the order of input elements within the node’s neighborhood.
\end{enumerate}

Despite these established forms of invariance, the exploration of invariance in graphs is still limited. Current research is primarily preliminary, focusing on aspects such as generalization bounds \cite{gargGeneralizationRepresentationalLimits2020a, vermaStabilityGeneralizationGraph2019} and permutation-invariant linear layers \cite{maronInvariantEquivariantGraph2019}, with few advances beyond these initial investigations.

Current research on graph invariance learning techniques can be categorized into two main areas \cite{suiUnleashingPowerGraph2023}.

\textbf{1. Invariant Graph Learning} focuses on capturing stable features by minimizing empirical risks \cite{chenDoesInvariantGraph2023}. 

\textbf{2. Graph Data Augmentation} encompasses both random and non-random methods, as detailed below:
\begin{itemize}
    \item \textbf{Random augmentation} introduces variability into graph features to improve generalization \cite{youGraphContrastiveLearning2020} and may include adversarial strategies \cite{sureshAdversarialGraphAugmentation2021}. However, excessive random augmentation can disrupt stable features and lead to uncontrolled distributions.
    \item \textbf{Non-Random Augmentation} involves specifically designed techniques such as graph rewiring \cite{sunBreakingEntanglementHomophily2023} and graph reduction \cite{hashemiComprehensiveSurveyGraph2024}. Graph reduction creates various perspectives of a graph through reductions at different ratios, thus augmenting the data for subsequent models. Examples include graph pooling \cite{gaoGraphPoolingNode2020}, multi-scale graph coarsening \cite{liangMILEMultiLevelFramework2021}, and using synthetic nodes to represent communities \cite{gaoMultipleSparseGraphs2023}. Among these, augmenting connections with high-order neighborhoods is a particularly popular technique.
    
\end{itemize}
 
\subsection{Higher-Order Neighborhood Aggregation}

Several methods extend GNNs by aggregating information from higher-order neighborhoods. These methods generally fall into two categories:
\begin{itemize}
    \item \textbf{Type 1: Powers of the Adjacency Matrix}  
This approach uses powers of the adjacency matrix \( A^k \). For example, MixHop \cite{abu-el-haijaMixHopHigherOrderGraph2019} aggregates messages from multi-hop neighbors by mixing different powers of the adjacency matrix. Adaptive Diffusions \cite{berberidisAdaptiveDiffusionsScalable2019} enhances this aggregation by sparsifying the matrix based on the landing probabilities of multi-hop neighbors. GPR-GNN \cite{chienAdaptiveUniversalGeneralized2021} introduces learnable weights for features from various orders, while H2GCN \cite{zhuHomophilyGraphNeural2020} combines MixHop with other techniques to address disassortative graphs. Additionally, \citet{zhangDiscoveringInvariantNeighborhood2024} investigates Invariant Neighborhood Patterns to manage shifts in neighborhood distribution, integrating both high-order and low-order information.

\item \textbf{Type 2: \( k \)-th Order Proximity}  
This method involves \( k \)-th order proximity, utilizing the multiplication of powers of the adjacency matrix $A$ with its transpose $A_t$: \( A^{k-1}A_t^{k-1} \). Techniques such as DiGCN(ib) \cite{tongDigraphInceptionConvolutional2020} and SymDiGCN \cite{tongDirectedGraphConvolutional2020} use this approach to capture richer neighborhood information.
\end{itemize}


To the best of our knowledge, there is currently no graph data augmentation method based on invariance.

%% file: 1_1_preliminary.tex
\section{Preliminaries}

Let \( G = (V, E) \) be a directed graph with \( n \) nodes and \( m \) edges, represented by an adjacency matrix \( A \in \{0,1\}^{n \times n} \), where \( A_{ij} = 1 \) indicates the presence of a directed edge from node \( i \) to node \( j \), and \( A_{ij} = 0 \) indicates the absence of such an edge. 
We focus on node classification where node features are organized in an \( n \times d \) matrix \( X \), where $d$ is dimension of features and node labels are \( y_i \in \{1, \ldots, C\} \). 
\begin{definition}[In-Neighbour]
     An \emph{in-neighbour} of a node \( v \in V \) is a node \( u \in V \) such that there is a directed edge from \( u \) to \( v \), i.e., \( (u, v) \in E \).
\end{definition}

\begin{definition}[Out-Neighbour]
     An \emph{out-neighbour} of a node \( v \in V \) is a node \( u \in V \) such that there is a directed edge from \( v \) to \( u \), i.e., \( (v, u) \in E \).
\end{definition}

\subsection{Scaled Ego-Graphs}
An $\alpha$-depth ego-graph \cite{alvarez-gonzalezWeisfeilerLehmanLocal2023} includes all nodes within $\alpha$ hops from a central node. We extend this concept to directed graphs and introduce scaled hops, leading to scaled ego-graphs.

\begin{definition}
In a directed graph \( G = (V, E) \), we define two types of \(\alpha\)-depth ego-graphs centered at a node \( v \in V \).

\begin{itemize}
    \item \textbf{\(\alpha\)-depth in-edge ego-graph}: \( I_\alpha(v) = (V_\leftarrow, E_\leftarrow) \), 
    
    where \( V_\leftarrow \) consists of all nodes that can reach \( v \) within \(\alpha\) steps, and \( E_\leftarrow \) consists of all directed edges between nodes in \( V_\leftarrow \) that are within \(\alpha\) steps of \( v \).
    
    \item \textbf{\(\alpha\)-depth out-edge ego-graph}: \( O_\alpha(v) = (V_\rightarrow, E_\rightarrow) \),
    
    where \( V_\rightarrow \) consists of all nodes that can be reached from \( v \) within \(\alpha\) steps, and \( E_\rightarrow \) consists of all directed edges from \( v \) to nodes in \( V_\rightarrow \) within \(\alpha\) steps.
\end{itemize}
\end{definition}

\begin{figure}[htbp]
    \centering
    \includegraphics[width=0.6\linewidth]{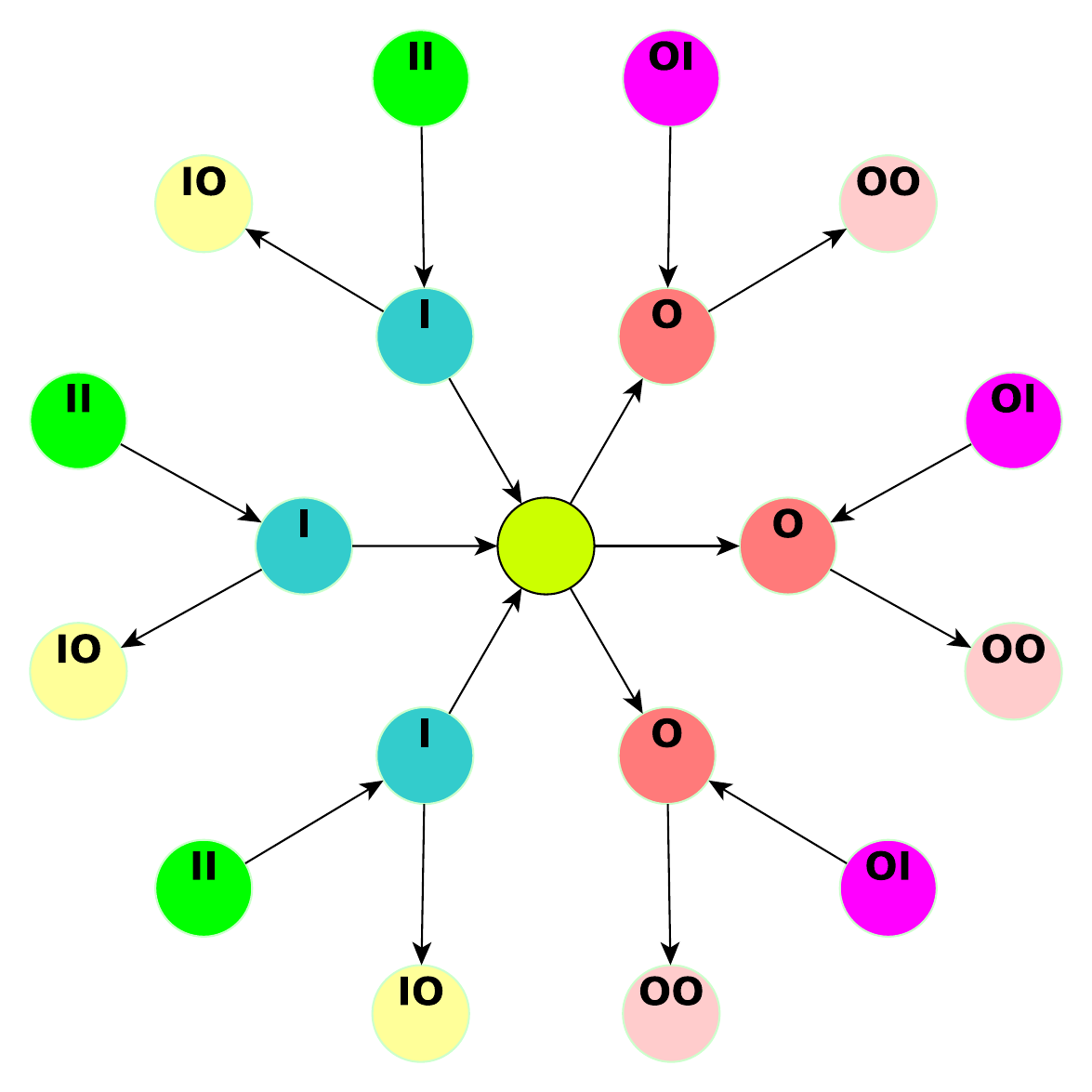}
 \caption{An illustration of scaled ego-graphs. For directed graphs, the 1-depth in-edge ego-graph comprises nodes labeled ``I'' along with the center node and all in-edges between them, whereas the 1-depth out-edge ego-graph comprises nodes labeled O along with the center node and all out-edges between them. The four types of 1-depth $2^{nd}$-scaled ego-graphs are composed of nodes labeled ``IO'', ``OI'', ``II'', and ``OO'', with the center node and all corresponding $2^{nd}$-scaled edges between them.}
    \label{fig:scaled_ego_graph}
\end{figure}

As illustrated in Figure \ref{fig:scaled_ego_graph}, a 1-depth ego-graph for an undirected graph includes nodes labeled I (in-neighbor) and O (out-neighbor). In the case of a directed graph, the 1-depth in-edge ego-graph comprises nodes labeled I along with the center node and all the edges connecting them, whereas the 1-depth out-edge ego-graph comprises nodes labeled O along with the center node and all the edges connecting them.

\begin{definition}
A \textbf{scaled-edge} is defined as an ordered sequence of multiple directed edges, where the \textbf{scale} refers to the number of edges in this sequence. Specifically, a \textbf{$k^{th}$-scale edge} is a scaled-edge composed of $k$ directed edges, also referred to as a \textbf{$k$-order edge}.

An \textbf{$\alpha$-depth scaled ego-graph} includes all nodes that are reachable within $\alpha$ hops of scaled-edge from a given center node.
\end{definition}
A $1^{st}$-scale edge, includes in-edge (I) and out-edge (O), connecting to in-neighbor (I) and out-neighbor (O) nodes, respectively, as shown in Figure \ref{fig:scaled_ego_graph}. Considering a $2^{nd}$-scale edge, there are four types: II, IO, OI, and OO, each connecting to nodes labeled in Figure \ref{fig:scaled_ego_graph} accordingly.

\subsubsection{Scale Invariance of Graphs}

The concept of scale invariance, well-known in image classification as the ability to recognize objects regardless of their size, can be extended to graphs. In the context of node classification, each node to be classified can be viewed as the center of an ego-graph. Thus, for node-level prediction tasks on graphs, each input instance is essentially an ego-graph $G_v$ centered at node $v$, with a corresponding target label $y_v$. Scale invariance in graphs would imply that the classification of a node remains consistent across different scaled ego-graphs.

\begin{definition}
Let $S_k$ denote the set of all $k^{th}$-scale edges and $G^k_\alpha(v)$ denote the set of all $\alpha$-depth $k^{th}$-scale ego-graphs centered at node $v$. Then we have the following equations:
\begin{equation}
\label{eq:sk}
    S_k = \{ e_1e_2\ldots e_k \mid e_i \in \{ \rightarrow, \leftarrow \}, 1 \leq i \leq k \} 
\end{equation}
\begin{equation}
\label{eq:Sk_alpha}
    G^k_\alpha(v) = \{(V_{s}, E_{s})  \mid s \in  S_k \}, 
\end{equation}
where \( e_1 e_2 \ldots e_k \) represents the scaled-edge obtained by following an ordered sequence of in-edge (\(\leftarrow\)) or out-edge (\(\rightarrow\)) hops from \( v \). Specifically:
\begin{itemize}
    \item \( V_s \) consists of all nodes that can be reached from \( v \) within \(\alpha\) steps of scaled-edge \( s \).
    \item \( E_s \) consists of all scaled-edges \( s \) from \( v \) to these nodes within those \(\alpha\) steps.
\end{itemize}
\end{definition}
\input{table/individual_scale}
Consider a GNN model \( M \) that learns from a graph \( G \) using its adjacency matrix \( A \) by aggregating information solely from its out-neighbors. To also learn from the in-neighbors, the model should aggregate information from the transpose of the adjacency matrix, i.e., \( A^T \) \cite{rossiEdgeDirectionalityImproves2023}.

An adjacency matrix which encodes scaled-edges is the ordered sequencial multiplication of \( A \) and \( A^T \).  The graph whose structure is represented with the scaled adjacency matrix is a scaled graph.
\begin{definition}[Scaled Adjacency Matrix and Scaled Graph]
Let $A_k$ denote the set of all $k^{th}$-scale adjacency matrix and $G^k$ denote the set of all $k^{th}$-scale graphs.
 \begin{equation}
A_k = \{ a_1a_2\ldots a_k \mid a_i \in \{ A, A^T \}, 1 \leq i \leq k \} 
 \end{equation}
 \begin{equation}
     G^k = \{ G^k = (V, \tilde{E_s}) \mid s \in  S_k \}, 
 \end{equation}
 where $\tilde{E_s}$ consists of all edges from the start of $s$ to the end of $s$. 
   
\end{definition}

To capture information from $2^{nd}$-scale neighbors, the model should extend its learning to matrices that incorporate both direct and transitive relationships. This involves using matrices such as \( AA \), \( AA^T \), \( A^TA^T \), and \( A^TA \) as the scaled adjacency matrix. 

\begin{definition}
For a node classification task on a graph $G$, we say the task exhibits scale invariance if the classification of a node $v$ remains invariant across different scales of its ego-graphs. Formally, for any $k \geq 1$:
\begin{equation}
    f(G_v) = f(G^k(v)),
\end{equation}
where $f$ is the classification function producing discrete values, $G_v$ is the original ego-graph of node $v$, and $G^k(v)$ is any $k^{th}$-scale ego-graph centered at $v$.
\end{definition}

This property implies that the essential structural information for node classification is preserved across different scales of the ego-graph. In other words, the $k^{th}$-scaled ego-graphs should maintain the node classification invariant.

\subsection{Demonstration of Scale Invariance}
\label{hetero_direction}

\input{table/chame_squi_neighbour}
\FloatBarrier
In Table \ref{tab:individual_scale}, we demonstrate the presence of scale invariance in graphs through various experiments. We represent the graph structure using a scaled adjacency matrix, which is then fed into a GNN for node classification. The results in Table \ref{tab:individual_scale} show that higher-scale graphs consistently achieve performance comparable to their lower-scale counterparts, confirming scale invariance. In contrast, if scale invariance were absent, the performance of higher-scale graphs would be similar to the results shown in the last column, where no input is provided.

Additionally, combinations of ego-graphs with adverse directional edges tend to yield better results than the individual ego-graphs. For example, $AA^T + A^TA$ generally achieves better accuracy compared to $AA^T$ and $A^TA$ individually, and $AA + A^TA^T$ performs better than $AA$ and $A^TA^T$.

For heterophilic graphs like Chameleon and Squirrel:
\begin{itemize}
    \item A outperforms $A^T$ in classification tasks. This means aggregating information from out-neighbors works better than from in-neighbors for these datasets.
    \item This trend extends to higher-order relationships: $AA^T$ and AA perform better than $A^TA$ and $A^TA^T$. This suggests mutual out-neighbors or 2-hop out-neighbors capture similarity more effectively than their in-neighbor counterparts.
    \item The reason: Most nodes have 0 neighbors when using $A^T$. After aggregation, these nodes' features are updated to all zeros.
\end{itemize}

Overall, Table \ref{tab:individual_scale} demonstrates that higher-scale graphs consistently perform no worse than their lower-scale counterparts, confirming the scale invariance of graph structures. The table also provides insights into how performance varies with different graph scales and characteristics across datasets.
\begin{figure}[ht]
    \centering
\includegraphics[width=\linewidth]{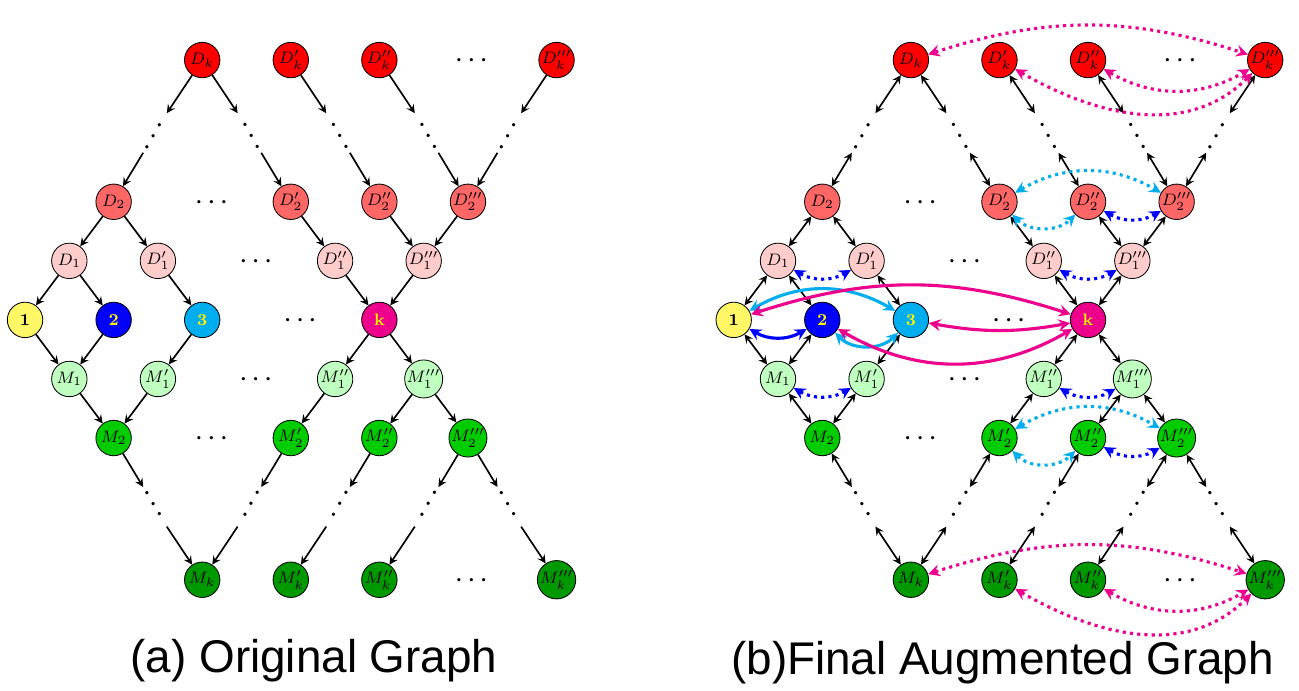}
    \caption{Edge augmentation by stacking multi-scale graphs in Digraph Inception Model. Details are provided in Appendix \ref{DiG<RiG}.}
    \label{fig:edge_aug}
\end{figure}

%% file: table/individual_scale.tex
\begin{table*}[ht]
   \centering
   \fontsize{9}{11}\selectfont
{
   \begin{tabular}{c|c|ccc|ccc|ccc|c}
   \toprule
   Type &   Dataset & $A$ & $A^T$ & $A$+$A^T$ & 
  $AA^T$ & $A^TA$ &  $AA$+$A^TA$ &
      $AA$  & $A^TA^T$ & $AA$+$A^TA^T$ & None \\
      \midrule
    \multirow{5}{*}{Homophilic}   & Telegram & 68 & 74 & 100 & 
        60 (58) & 68 (78) &  92 (94) & 72 (50)  & 80 (78) & 92 (92)  & 38 \\
    &  Cora-ML & 75 & 70 & 78 & 
      78 (73) & 77 (69) &  79 (78) &
      75 (80)  & 72 (78) & 77 (79) & 29 \\
   & CiteSeer & 56 & 59 & 62 & 
      62 (57) & 64 (59) &  63 (61) &
      57 (53)  & 60 (60) & 63 (63) & 20 \\
      
      
   &   WikiCS & 73 & 66 & 75 & 
      74 (76) & 65 (69) &  76 (78) &
       73 (75)  & 66 (67) & 73 (77) & 23 \\
   
    \midrule
    \multirow{2}{*}{Heterophilic}     & Chameleon & 78 & 30 & 68 & 
      66 (70) & 29 (29) &  68 (71) &
      70 (76)  & 30 (31) & 66 (69) & 22 \\
 &      Squirrel & 75 & 33 & 68 & 
      73 (73) & 31 (31) &  70 (67) &
      75 (73)  & 32 (33) & 66 (66) &  19 \\
 \bottomrule
    \end{tabular}}

   
    \caption{Accuracy obtained by each scaled ego-graph using a single split per ego-graph. 
 Higher scaled graphs maintain the discerning ability of their lower scale counterparts, even after removing the shared edges from lower scale graphs (in parentheses). For homophilic graphs, both \(A\) and \(A^T\) perform well, and all $2^{nd}$-scale graphs preserve this performance. In heterophilic graphs, \(A\) performs well while \(A^T\) does not. \(AA\) and \(AA^T\) preserve the performance of \(A\), whereas \(A^TA^T\) and \(A^TA\) inherit the poor performance of \(A^T\). (The splits are shown in Appendix \ref{data_appendix}. The final column presents the performance with all zero input for comparison. The value in parentheses is the scaled graph after removing the shared edges with $A$ or $A^T$. ``+'' denotes the addition of aggregation outputs.)}
    \label{tab:individual_scale}
\end{table*}

%% file: table/chame_squi_neighbour.tex
\begin{table}[ht]
    \centering
{
    \begin{tabular}{cc|ccc}
    \toprule
    Dataset &  Direction & Homo. & Hetero. & No Neigh. \\
    \midrule
    \multirow{2}{*}{Chameleon} 
    & $A$  & 576  & 1701  & 0    \\
     & $A^T$    & 237  & 627   & 1413 \\
     
    \midrule
    \multirow{2}{*}{Squirrel}  
    & $A$  & 1258 & 3943  & 0    \\
     & $A^T$    & 441  & 1764  & 2996 \\
     
    \bottomrule     
    \end{tabular}}
\caption{This figure illustrates the distribution of nodes in the Chameleon and Squirrel datasets, categorized by the predominant label of their aggregated neighbors in relation to the node's own label. The analysis reveals that when using $A^T$ as the adjacency matrix, a majority of nodes have zero aggregated neighbors. This lack of connectivity results in poor model performance}
    \label{tab:explain_dir for chame}
\end{table}

%% file: 1_2_methodology.tex
\section{ScaleNet}

Given the observed scale invariance across all tested datasets, combining these scaled graphs might further enhance the performance of GNNs. Digraph Inception Networks are one approach to achieving this.
\subsection{Digraph Inception Networks}
\label{inception}

State-of-the-art Graph Neural Networks (GNNs) for homophilic graphs include Graph Inception Networks such as DiGCN(ib) \footnote{In this paper, DiGCN is also called DiG, and DiGCNib is also called DiGib, DiGi2.}  \cite{tongDigraphInceptionConvolutional2020} and SymDiGCN \footnote{In this paper, SymDiGCN is also called Sym for short.} \cite{tongDirectedGraphConvolutional2020}, which incorporate higher-order proximity to obtain multi-scale features. However, these methods are based on random walks, making edge weights across various scales crucial. DiGCN(ib) uses resource-intensive eigenvalue decomposition to determine weights, while SymDiGCN relies on costly incidence normalization based on node degrees in the original graph, which limits their scalability to larger graphs.
\FloatBarrier
The essence of their success is the equivalent edge augmentation, as shown in Figure \ref{fig:edge_aug}.
We revised their models by removing the computationally expensive edge weight calculation block and replacing it with a simple constant weight of \textbf{1} (For specific details, please refer to Appendix \ref{inception_Appendix}). The results, shown in the ``Symmetric Type'' section of Table  \ref{tab:compare} indicate that \textbf{1}iG generally outperforms or matches DiG across all datasets, with \textbf{1}iGi2 showing similar improvements over DiGib, and \textbf{1}ym over Sym (The meanings of \textbf{1}iG, \textbf{1}iGi2, \textbf{1}ym are explained in Table \ref{tab:compare}).

We further demonstrate that the heavily computed edge weights used by DiGCN \cite{tongDigraphInceptionConvolutional2020} are not always desirable, as they may result in worse performance compared to even randomly assigned weights.

In addition, as shown in Tables \ref{tab:compare} and \ref{tab:imbalanced}, Sym and DiG encounter out-of-memory issues on larger datasets, unlike the other models.

Overall, the existing digraph inception models are resource-intensive, and the costly edge weights they produce are not ideal. Our inception models achieve better performance by simply assigning a value of 1 to all scaled edges, making them simpler, faster, and more accurate.

\subsection{A Unified Network: ScaleNet}
\input{fig/scalenet}

As discussed in Section \ref{hetero_direction}, heterophilic graphs tend to suffer from performance degradation when aggregating information from scaled graphs in both directions. This limitation causes existing Digraph Inception Networks to perform poorly on heterophilic graphs.

To address this issue and accommodate the unique characteristics of different datasets, we propose a flexible combination approach and introduce \textbf{ScaleNet}, as illustrated in Figure \ref{fig:multilayerScaleNet}. This approach flexibly synthesizes scaled graphs and optionally integrates components like self-loops, batch normalization, and non-linear activation functions, each of which is tailored to the specific characteristics of the dataset through a grid search of model parameters.

\paragraph{Bidirectional Aggregation}

To exploit scale invariance, we define the bidirectional aggregation function \(\textbf{AGG-B}_\alpha(M, N, X)\) as follows:
\begin{equation}
\label{alpha_dir}
    (1+\alpha)\alpha \, \textbf{AGG}(M, X) + (1+\alpha)(1-\alpha) \, \textbf{AGG}(N, X)
\end{equation}
The \(\textbf{AGG}\) function can be any message-passing neural network (MPNN) architecture, such as  GCN \cite{kipfSemiSupervisedClassificationGraph2017}, GAT \cite{velickovicGraphAttentionNetworks2018}, or SAGE \cite{hamiltonInductiveRepresentationLearning2018}. 
$M$ and $N$ represent pairs of matrices encoding opposite directional edges. 
The parameter \(\alpha\) controls the contribution of matrices \(M\) and \(N\): \(\alpha = 0\) uses only \(M\), \(\alpha = 1\) uses only \(N\), \(\alpha = 0.5\) balances both, and \(\alpha = -1\) excludes both.

Given that adding or removing self-loops \cite{kipfSemiSupervisedClassificationGraph2017, tongDigraphInceptionConvolutional2020} can influence the performance of the model, we allow for the inclusion of such options by defining $\tilde{A}$, which can be: (i) the matrix $A$ with self-loops being removed, (ii) the matrix $A$ with self-loops being added, or (iii) the original matrix $A$. The influence of self-loops is shown in Appendix \ref{A+selfloop}.

This formulation provides a flexible framework for aggregating information from bidirectional matrices, enabling the model to leverage various directional and self-loop configurations to enhance its performance.

Additionally, setting \(\alpha = 2\) combines the matrices \(M\) and \(N\) directly before aggregation, while setting \(\alpha = 3\) considers their intersection:
\begin{equation}
    \textbf{AGG-B}_2(M, N, X) = \textbf{AGG}(M \cup N, X)
\end{equation}
\begin{equation}
    \textbf{AGG-B}_3(M, N, X) = \textbf{AGG}(M \cap N, X)
\end{equation}

\paragraph{Layer-wise Aggregation of ScaleNet}
We combine the propagation output from various scaled graphs with the following rule:
\begin{equation}
X^{(l)} = \textbf{COMB1}(X^{(l)}_{1}, X^{(l)}_{2}, X^{(l)}_{3}, \ldots),
\end{equation}
where \( X^{(l)} \) represents the updated features after \( l \) layers. The function \textbf{COMB1} can be realized by the Jumping Knowledge (JK) framework as proposed by Xu et al. \cite{xuRepresentationLearningGraphs2018}, or simply by performing an element-wise addition of the inputs. 
\paragraph{Multi-layer ScaleNet}
A multi-layer ScaleNet is then defined as follows:
\begin{equation}
    \textbf{Output} = \textbf{COMB2}(X^{(1)}, X^{(2)}, \ldots, X^{(L)})
\end{equation}
In this formulation, \( L \) layers of the propagation rule are stacked. The function \textbf{COMB2} combines the outputs of all layers, which can again be done using the Jumping Knowledge technique, or alternatively, the output from the final layer may be used directly as the model's output.

\section{Experiments}
\subsection{Robustness to Imbalanced Graphs}

ScaleNet improves robustness against imbalanced graphs by leveraging multi-scale graphs, similar to data augmentation techniques.

Table \ref{tab:imbalanced} indicates that ScaleNet consistently outperforms Dir-GNN and MagNet on imbalanced datasets. The imbalance ratio measures the size disparity between the largest and smallest classes. ScaleNet’s advantage stems from its use of higher-scale graphs and self-loops, which enhances its ability to capture essential features in homophilic graphs that Dir-GNN and MagNet might miss. Conversely, single-scale networks like Dir-GNN \cite{rossiEdgeDirectionalityImproves2023} and MagNet \cite{zhangMagNetNeuralNetwork} are prone to over-smoothing and may incorporate irrelevant nodes due to excessive layer stacking when aggregating information from longer-range nodes.
\input{table/imbalan}

\subsection{Performance of ScaleNet on Different Graphs}
\input{table/total_compare}
\input{table/wilcox_all}

We present the 10-split cross-validation experimental results in Table \ref{tab:compare} and compare the top two models using the Wilcoxon signed-rank test with 30 splits, as shown in Table \ref{tab:wilcoxon_all}. For a detailed analysis of the Wilcoxon signed-rank test applied to additional models, please refer to Appendix \ref{appen_wilcoxon}.

ScaleNet achieves the best performance on all homophilic graphs, with our symmetric models securing the second-best performance. 

For heterophilic graphs, ScaleNet matches the performance of Dir-GNN \cite{rossiEdgeDirectionalityImproves2023}. The Wilcoxon signed-rank test shows that ScaleNet significantly outperforms Dir-GNN on Chameleon, with a p-value less than 0.05, indicating that the observed difference is statistically significant. On Squirrel, however, the performance of both models is similar.

Overall, ScaleNet is the top-performing model across all datasets, significantly outperforming existing models on 5 datasets and matching the best on 2 datasets.

\subsection{Tuning ScaleNet for Different Graph Types}

ScaleNet is designed to adapt to the unique characteristics of each dataset, offering optimal performance for both homophilic and heterophilic graphs.

During tuning with a grid search of hyperparameters, we observed the following findings:

\begin{itemize}
    \item Homophilic Graphs: performance is improved by adding self-loops and using both scaled graphs based on opposite directed scaled edges, such as $AA$ and $A^TA^T$.
    \item Heterophilic Graphs: performance benefits from removing self-loops and utilizing scaled graphs with preferred directional scaled edges, while excluding those based on the opposite directional scaled edges.
    \item Additional findings:
\begin{itemize}
    \item For imbalanced datasets such as the Telegram, incorporating batch normalization significantly improves performance.
\item The CiteSeer dataset  performs better with the removal of nonlinear activation functions.
\end{itemize}
\end{itemize}

Our unified model, optimized through grid search, reveals the characteristics of different graph datasets and provides a strong basis for model comparison.

%% file: fig/scalenet.tex
\begin{figure}[ht]
    \centering
\includegraphics[width=\linewidth]{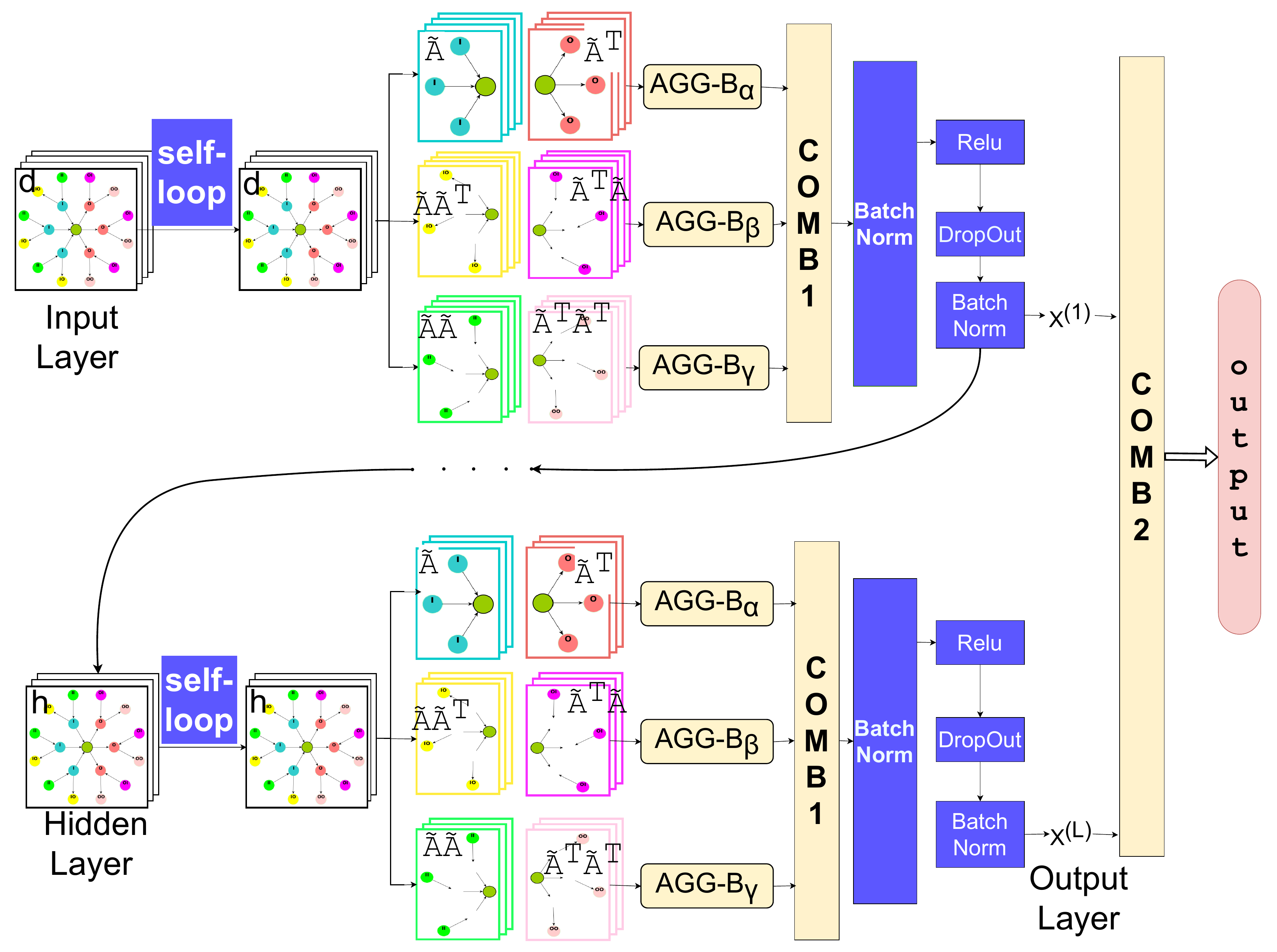 }
\caption{Schematic depiction of multi-layer ScaleNet with d input channels and h hidden channels. 
For layer-wise aggregation, the original graph is derived into two $1^{st}$-scaled and four $2^{nd}$-scaled graphs. Three \textbf{AGG-B} blocks determine input selection for \textbf{COMB1}, which uses either a jumping knowledge architecture or addition. \textbf{COMB2} represents the fusion of all layers' outputs. (The blue blocks are optional, including self-loop operations, non-linear activation functions, dropout, and layer normalization.)}
    \label{fig:multilayerScaleNet}
\end{figure}

%% file: table/imbalan.tex
\begin{table}[ht]
    \centering
\setlength{\tabcolsep}{0.5mm}
{
   \begin{tabular}{c c |ccccc|cc}
   \toprule
 Type &  Method & Cora-ML & CiteSeer
 & WikiCS \\
 \midrule
 \multirow{2}{*}{Standard} 
  & MagNet & 47.9±5.5 & 29.3 
  & 62.0±1.5 \\
  
& Dir-GNN  & 41.1 & 25.0 
& 62.9±1.4  \\

\midrule
 \multirow{5}{*}{Augment} 
& DiG & 60.9±1.8 & 36.9 
& 72.2±1.4  \\
& DiGib  & 55.7±2.9 & 40.4 
& 69.8±1.2  \\

& \textbf{1}iG  & 64.9±4.7 & 42.3 
& 71.0±1.5  \\
& \textbf{1}iGi2 & 61.9±5.7 & 41.5 
& 71.0±1.6 \\

& ScaleNet  & 60.3±6.7 & 43.1 
& 69.4±1.2  \\
 \bottomrule     
    \end{tabular}}
    \caption{Accuracy on imbalanced datasets (imbalance ratio = 100:1). When accuracy is below 45\%, only one split is used. 
    }
    \label{tab:imbalanced}
\end{table}

%% file: table/total_compare.tex
\begin{table*}[ht]
        \centering

{
   \begin{tabular}{cc|cccc|cc}
   \toprule
 Type & Method & Telegram & Cora-ML & CiteSeer 
 & WikiCS & Chameleon & Squirrel \\
 \midrule
\multirow{5}{*}{Base models} 
& MLP & 32.8±5.4 & 67.3±2.3 & 54.5±2.3 
& 73.4±0.6 & 40.3±5.8 & 28.7±4.0 \\

& GCN & 86.0±4.5 & 81.2±1.4 & 65.8±2.3 
& 78.8±0.4 & 64.8±2.2 & 46.3±1.9  \\
& APPNP &  67.3±3.0 & 81.8±1.3 & 65.9±1.6 
& 77.6±0.6 & 38.7±2.4 & 27.0±1.5 \\
& ChebNet & 83.0±3.8 & 80.5±1.6 & 66.5±1.8 
& 76.9±0.9 & 58.3±2.4 & 38.5±1.4 \\
& SAGE & 74.0±7.0 & 81.7±1.2 & 66.7±1.7 
& \textbf{79.3±0.4} & 63.4±3.0 & 44.6±1.3 \\

 \midrule
\multirow{3}{*}{Hermitian} & MagNet & 87.6±2.9 & 79.7±2.3 & 66.5±2.0 
& 74.7±0.6 & 58.2±2.9 &  39.0±1.9\\
& SigMaNet &  86.9±6.2 & 71.7±3.3 & 44.9±3.1 
& 71.4±0.7 & 64.1±1.6 & OOM  \\
& QuaNet &  85.6±6.0 & 26.3±3.5 & 30.2±3.0 
& 55.2±1.9 & 38.8±2.9 & OOM  \\
\midrule
 \multirow{3}{*}{Symmetric} 
& Sym & 87.2±3.7 & 81.9±1.6 & 65.8±2.3 
& OOM & 57.8±3.0 & 38.1±1.4 \\
 & DiG & 82.0±3.1 & 78.4±0.9 & 63.8±2.0 
 & 77.1±1.0 & 50.4±2.1 & 39.2±1.8 \\
 & DiGib & 64.1±7.0 & 77.5±1.9 & 60.3±1.5 
 & 78.3±0.7 & 52.2±3.7 & 37.7±1.5\\

 \midrule
 \multirow{3}{*}{\makecell{Symmetric\\(\textbf{Ours})}}
 &  \textbf{1}ym & 84.0±3.9 & 80.8±1.6 & 64.9±2.5 
 & 75.4±0.4 & 54.9±2.7 & 35.5±1.1 \\
 & \textbf{1}iG & \underline{95.8±3.5} & 82.0±1.3 & 65.5±2.4 
 & 77.4±0.6 & 70.2±1.6 & 50.7±5.8 \\
& \textbf{1}iGi2 & 93.0±5.1 & 81.7±1.3 & \underline{67.9±2.2}
&   78.9±0.6 & 58.4±2.5 & 42.7±2.5 \\
  & \textbf{1}iGu2 & 92.6±4.9 & \underline{82.1±1.2} & 67.6±1.8 
  & 75.6±0.9 & 60.4±2.4 & 40.4±1.8 \\

\midrule
  
 BiDirection & Dir-GNN & 90.2±4.8 & 79.2±2.1 & 61.6±2.6 
 & 77.2±0.8 & \underline{79.4±1.4} & \underline{75.2±1.7}  \\
\midrule
 \multirow{2}{*}{\textbf{Ours}} & ScaleNet &\textbf{96.8±2.1} & \textbf{82.3±1.1} & \textbf{68.3±1.5} 
 & \textbf{79.3±0.6} & \textbf{79.6±1.2} & \textbf{75.3±2.0}    \\
 & $\alpha, \beta, \gamma$ &  0.5,-1,-1 & 2,-1,-1 & 0.5,2,-1 
 & 0.5,2,-1 & 1,1,1 & 1,1,1  \\
 
   \bottomrule
    \end{tabular}}
\caption{Node classification Accuracy (\%). The best results are in \textbf{bold} and the second best are \underline{underlined}. 
10-fold cross validation is used. OOM indicates out of memory on GPU3090 with 24GB of VRAM. Model \textbf{1}ym assigns 1 to edge weights of model Sym: similarly, model \textbf{1}iG and \textbf{1}iGi2 are assigning 1 to edge weights of models DiG and DiGib, respectively. Model \textbf{1}iGu2 and \textbf{1}iGu3 assign weights of 1 to scaled edges, but use union instead of intersection in DiG\textbf{i}b, and the last number k denotes the model includes up to $k^{th}$-scale edges, while DiGib only scales up to $2^{nd}$-order. At the end of model name, ``ib'' would be used interchangeably with ``i2''. Parameters $\alpha$, $\beta$, and $\gamma$ controlling ScaleNet components: $\alpha$ controls $A$ and $A^T$, $\beta$ controls $AA^T$ and $A^TA$, and $\gamma$ controls $AA$ and $A^TA^T$.
}

    \label{tab:compare}
\end{table*}

%% file: table/wilcox_all.tex
\begin{table*}[ht]
    \centering
{
    \begin{tabular}{l|c|c|c|c|c|c}
    \toprule
     & Telegram & Cora-ML & CiteSeer 
    & WikiCS & Chameleon & Squirrel    \\
     \midrule
     Methods &  \makecell{ScaleNet \\ \textbf{1}iG}    
 & \makecell{ScaleNet \\ \textbf{1}iGu2}  
 & \makecell{ScaleNet \\ \textbf{1}iGi2}

     
     &  \makecell{ScaleNet \\ SAGE} 
     &  \makecell{ScaleNet \\ Dir-GNN}
     &  \makecell{ScaleNet \\ Dir-GNN}\\
     \midrule
\makecell{p-value \\ (Statistic)}
& \makecell{\textbf{0.0064} \\ (62.0)} 
& \makecell{0.1241 \\ (156.5)} 
& \makecell{\textbf{0.0013} \\ (82.0)} 


& \makecell{0.9515 \\(229.0) } 
& \makecell{\textbf{0.0095} \\ (97.5)}
& \makecell{0.3085 \\(181.5)}\\
\midrule
Best Model
& ScaleNet & Both & ScaleNet  & Both 
& ScaleNet & Both\\

\bottomrule
    \end{tabular}}
\caption{Results of the Wilcoxon signed-rank test for the top two models, based on 30 splits. Statistical significance is indicated by p-values less than 0.05, with significant p-values shown in bold. If p-value is not below 0.05, both models are considered equally effective.}

    \label{tab:wilcoxon_all}
\end{table*}

%% file: conclusions.tex
\section{Conclusions}

This research provides compelling evidence of scale invariance in graphs, leading to the development of simplified inception models that achieve state-of-the-art performance on homophilic datasets. To address the limitations of these inception networks on heterophilic graphs, we introduce ScaleNet—a unified model that excels in both homophilic and heterophilic settings. ScaleNet achieves this by flexibly integrating scaled graphs and optional components such as batch normalization and non-linear activation. To the best of our knowledge, this is the first model to achieve state-of-the-art performance across both types of graphs.
\FloatBarrier

%% file: appendix_data.tex
\input{table/statistics_data}
\section{More details on experiments}
\label{data_appendix}
\subsection{Datasets}

We use six widely-adopted real-world datasets, comprising four homophilic and two heterophilic graph datasets. To ensure consistency and comparability, we maintain the original train/validation/test splits provided by the source datasets. 
All datasets have 10 splits, except WikiCS, which originally includes 20 splits.

CiteSeer, Cora-ML, and WikiCS are citation networks, while Telegram is a social network. Chameleon and Squirrel are webpage networks. Dataset statistics have been reported in Table \ref{tab:statis}.
\begin{itemize}
\item \textbf{Homophilic Graph Datasets}
    \begin{itemize}
    \item Telegram is characterized as a network comprising pairwise interactions between various elements, including channels, chats, posts, and URL links within the platform. 
    
    We follow the splits described in MagNet \cite{zhangMagNetNeuralNetwork}.
    \item For citation networks, nodes represent documents and papers, while edges denote citation links. The classes correspond to different research domains.
    \begin{itemize}
        \item For Citeseer and Cora-ML, we use the splits specified in the DiGCN(ib) paper \cite{tongDigraphInceptionConvolutional2020}.
        \item For the WikiCS dataset, the splits are described in \cite{mernyeiWikiCSWikipediaBasedBenchmark2022}.
    \end{itemize}.
    \end{itemize}

\item \textbf{Heterophilic Graph Datasets}
    \begin{itemize}
        \item Chameleon and Squirrel are Wikipedia page networks focused on specific topics, with nodes representing web pages and edges representing links between them \cite{rozemberczki2021multi}. Nodes have features based on important nouns from the Wikipedia pages and are classified into five categories based on their average monthly traffic. 
        
        These classifications follow the splits used in GEOM-GCN \cite{peiGeomGCNGeometricGraph2020}.        
    \end{itemize}

\end{itemize}

\subsection{Experimental Settings}
\subsubsection{Implementation Details}
All experiments use a GPU3090 with 24GB of memory, except for the experiments on imbalanced datasets shown in Table 3 
of the main text, which were conducted on a GPU4070 with 12GB of memory.

For all experiments, we used the Adam optimizer and trained the model for 1,500 epochs, applying early stopping based on validation accuracy with a patience of 410 epochs. A learning rate scheduler with a patience of 80 epochs was also used for all datasets.

We report the mean and standard deviation of test accuracy over 10 runs, except for WikiCS, where the results are based on 20 splits as originally provided.

\subsubsection{Hyperparameter Tuning}

For each model, we perform a grid search to optimize the following hyperparameters:

\begin{itemize}
    \item \textbf{Number of Layers:} from 1 to 5
    \item \textbf{Learning Rate:} 0.1, 0.01, 0.005
    \item \textbf{Dropout Rate:} 0.0, 0.5
    \item \textbf{batch normalization:} Enabled(1) or Disabled(0)
    \item \textbf{ReLU Activation:} Applied(1) or Not Applied(0)
    \item \textbf{Jumping Knowledge (JK) Aggregation:} \texttt{max}, \texttt{cat}, or \texttt{0} (no jumping knowledge structure)
    \item \textbf{Self-loop Handling:} Added (add), Removed (remove), or None (0)
    \item \textbf{Directional Parameter} in Equation 6
    : 0, 0.5, 1, 2, 3
\end{itemize}

We consider both options for JK structures, batch normalization, and ReLU Activation, applied either within layer-wise ScaleNet or across multiple layers of ScaleNet.

For higher scaled adjacency matrices, such as $AA^T$ and $A^TA$, there are options to either remove or retain generated self-loops.

\subsection{Wilcoxon Signed-Rank Test Analysis}
\label{appen_wilcoxon}

We conducted Wilcoxon signed-rank tests to compare the performance of the top models across seven different datasets. The results are presented in Tables \ref{tab:wilcoxon_tel} to \ref{tab:wilcoxon_squir}. Each table displays the mean performance ± standard deviation for each model on the diagonal, with p-values and test statistics for pairwise comparisons on the off-diagonal. P-values less than 0.05 are bolded to indicate statistically significant differences.

\input{table/wilcoxon/wilcoxon_tel}

\input{table/wilcoxon/wilcoxon_cora}

\input{table/wilcoxon/wilcoxon_cite}

\input{table/wilcoxon/wilcoxon_wikics}

\input{table/wilcoxon/wilcoxon_chameleon}

\input{table/wilcoxon/wilcoxon_squirrel}


%% file: table/statistics_data.tex
\begin{table*}[ht]
    \centering
   \fontsize{9}{11}\selectfont
{
    \begin{tabular}{c|ccccc|ccccc}
    \toprule
Dataset & \#Nodes & \#Edges &\#Feat. &  \#C & Imbal-Ratio & \%No-In & \%In-Homo & \%No-Out & \%Out-Homo  & Label rate\\
CiteSeer &3312 &4715 &3703 & 6 & 1.0 & 30.2 & 52.7 & 41.1 & 44.1 &  3.6\% \\
Cora-ML  & 2995&8416 & 2879&7 &1.0 &  41.7 & 50.2 & 11.7 & 74.7 &  4.8\% \\
Telegram & 245& 8912 & 1&4 & 3.0 &  16.7 & 32.2 & 25.3 & 32.7 &  60\% \\
PubMed &19717 &44327 &500 &3 &1.0 &  33.4 & 54.2 & 34.2 & 54.0 &  0.3\%\\
WikiCS  &11701 &297110 &300 &10 & 9.5 &  18.5 & 69.1 & 3.8 & 76.7  &5\% \\
\midrule
Chameleon & 2277 & 36101 & 2325& 5 & 1.3 & 62.1 & 10.4 & 0.0 & 25.3 & 48\%  \\
Squirrel &5201 & 217073&2089 & 5 & 1.1&  57.6 & 8.5 & 0.0 & 24.2 & 48\% \\
\bottomrule     

    \end{tabular}}
\caption{Dataset statistics. Imbalance Ratio is the ratio of the largest class to the smallest class in the training sets. \%No-In and \%No-Out represent the percentage of nodes with no direct in-neighbors and no direct out-neighbors, respectively. \%In-Homo denotes the percentage of nodes whose in-neighbors predominantly share the same label as the node, while \%Out-Homo indicates the percentage of nodes whose out-neighbors predominantly share the same label.}

    \label{tab:statis}
\end{table*}

%% file: table/wilcoxon/wilcoxon_tel.tex

\begin{table*}[ht]
    \centering
    \begin{tabular}{l|c|c|c|c}
        \toprule
  \textbf{Dataset: Telegram}      & ScaleNet & \textbf{1}iG & Dir-GNN & MagNet \\
        \midrule
        ScaleNet & 95.93 ± 2.99 & \textbf{0.0064} (62.0) & \textbf{0.0001} (18.0) & \textbf{0.0000} (0.0) \\
        \midrule
        \textbf{1}iG & \textbf{0.0064} (62.0) & 93.27 ± 4.30 & 0.3912 (131.0) & \textbf{0.0002} (35.5) \\
        \midrule
        Dir-GNN & \textbf{0.0001} (18.0) & 0.3912 (131.0) & 92.20 ± 3.66 & \textbf{0.0000} (20.0) \\
        \midrule
        MagNet & \textbf{0.0000} (0.0) & \textbf{0.0002} (35.5) & \textbf{0.0000} (20.0) & 86.60 ± 5.42 \\
        \bottomrule
    \end{tabular}
    \caption{Wilcoxon signed-rank test results for pairwise comparisons among the top models (ScaleNet, \textbf{1iG}, Dir-GNN, and MagNet) on the Telegram dataset. ScaleNet is identified as the superior model for
this dataset. The results indicate that ScaleNet
consistently outperforms the other models, demonstrating
statistically significant improvements in performance.}
    \label{tab:wilcoxon_tel}
\end{table*}

%% file: table/wilcoxon/wilcoxon_cora.tex
\begin{table*}[ht]
    \centering
    \begin{tabular}{l|c|c|c|c|c}
        \toprule
      \textbf{Dataset: Cora-ML}  & ScaleNet & \textbf{1}iGu2 & \textbf{1}iG & Sym & \textbf{1}ym \\
        \midrule
        ScaleNet & 82.22 ±1.16 & 0.3043 (170.0) & 0.1241 (156.5) & \textbf{0.0001} (56.0) & \textbf{0.0002} (60.5) \\
        \midrule
        \textbf{1}iGu2 & 0.3043 (170.0) & 82.43 ±1.48 & 0.1642 (163.5) & \textbf{0.0001} (53.5) & \textbf{0.0001} (41.0) \\
        \midrule
        \textbf{1}iG & 0.1241 (156.5) & 0.1642 (163.5) & 81.85 ±1.27 & \textbf{0.0099} (108.5) & \textbf{0.0076} (105.0) \\
        \midrule
        Sym & \textbf{0.0001} (56.0) & \textbf{0.0001} (53.5) & \textbf{0.0099} (108.5) & 80.75 ±1.84 & 0.6575 (197.0) \\
        \midrule
        \textbf{1}ym & \textbf{0.0002} (60.5) & \textbf{0.0001} (41.0) & \textbf{0.0076} (105.0) & 0.6575 (197.0) & 80.90 ±1.45 \\
        \bottomrule
    \end{tabular}
    \caption{Wilcoxon signed-rank test results for pairwise comparisons among the models ScaleNet, \textbf{1}iGu2, \textbf{1}iG, Sym, and \textbf{1}ym on the Cora-ML dataset. The best models for the
Cora-ML dataset are ScaleNet, 1iGu2, and 1iG, as they show
no significant differences.}
    \label{tab:wilcoxon_cora}
\end{table*}

%% file: table/wilcoxon/wilcoxon_cite.tex

\begin{table*}[ht]
    \centering
    \begin{tabular}{l|c|c|c|c}
        \toprule
     \textbf{Dataset: CiteSeer}   & ScaleNet & \textbf{1}iGi2 & Sym & \textbf{1}iGu2 \\
        \midrule
        ScaleNet 
        & 68.13 ±1.46 
        & \textbf{0.0013} (82.0) 
        & \textbf{0.0000} (3.0) 
        & \textbf{0.0000} (27.0) \\
        \midrule
        \textbf{1}iGi2 
        & \textbf{0.0013} (82.0) 
        & 66.98 ±2.09 
        & \textbf{0.0003} (65.0) 
        & 0.1981 (169.0) \\
       \midrule
        Sym 
        & \textbf{0.0000} (3.0) 
        & \textbf{0.0003} (65.0) 
        & 65.44 ±2.43 
        & 0.0767 (146.0) \\
       \midrule
        \textbf{1}iGu2 
        & \textbf{0.0000} (27.0) 
        & 0.1981 (169.0) 
        & 0.0767 (146.0) 
        & 66.22 ±1.61 \\
        \bottomrule
    \end{tabular}
    \caption{Wilcoxon signed-rank test results for pairwise comparisons among the top models (ScaleNet, \textbf{1}iGi2, Sym, and \textbf{1}iGu2) on the CiteSeer dataset. ScaleNet stands out as the
best model for the CiteSeer dataset.}
    \label{tab:wilcoxon_citeseer}
\end{table*}

%% file: table/wilcoxon/wilcoxon_wikics.tex
\begin{table*}[ht]
    \centering
\begin{tabular}{l|c|c|c|c|c}
    \toprule
    \textbf{Dataset: WikiCS} & ScaleNet & SAGE & \textbf{1}iGi2 & DiGib & Dir-GNN  \\
    \midrule
    ScaleNet & 79.30 ± 0.51 & 0.9515 (229.0) & \textbf{0.0000} (38.5) & \textbf{0.0000} (1.0) & \textbf{0.0000} (0.0) \\
    \midrule
    SAGE & 0.9515 (229.0) & 79.22 ± 0.50 & \textbf{0.0000} (25.0) & \textbf{0.0000} (0.0) & \textbf{0.0000} (0.0) \\
    \midrule
    \textbf{1}iGi2 & \textbf{0.0000} (38.5) & \textbf{0.0000} (25.0) & 78.52 ± 0.66 & \textbf{0.0000} (10.0) & \textbf{0.0000} (0.0) \\
    \midrule
    DiGib & \textbf{0.0000} (1.0) & \textbf{0.0000} (0.0) & \textbf{0.0000} (10.0) & 77.31 ± 0.73 & \textbf{0.0000} (1.0) \\
    \midrule
    Dir-GNN & \textbf{0.0000} (0.0) & \textbf{0.0000} (0.0) & \textbf{0.0000} (0.0) & \textbf{0.0000} (1.0) & 75.67 ± 0.75 \\
    \bottomrule
\end{tabular}
    
    \caption{Wilcoxon signed-rank test results for pairwise comparisons among the top models (ScaleNet, SAGE, \textbf{1iGi2}, DiGib, and Dir-GNN) on the WikiCS dataset. ScaleNet and SAGE are the best models, showing a significant difference from other models while not showing a significant difference from each other.}
    \label{tab:wilcoxon_wikics}
\end{table*}

%% file: table/wilcoxon/wilcoxon_chameleon.tex
\begin{table*}[ht]
    \centering
    \fontsize{9}{11}\selectfont
    \begin{tabular}{l|c|c|c|c|c}
       \toprule
  \textbf{Dataset: Chameleon}      & ScaleNet \((1,1,1)\) 
        & ScaleNet \((1,-1,1)\) 
        & ScaleNet \((1,-1,-1)\) 
        & Dir-GNN 
        & \textbf{1}iG \\
        \midrule
        ScaleNet \((1,1,1)\) 
        & 79.27 ±1.60 
        & \textbf{0.0000} (48.0) 
        & \textbf{0.0227} (103.0) 
        & \textbf{0.0095} (97.5) 
        & \textbf{0.0000} (0.0) \\
       \midrule
        ScaleNet \((1,-1,1)\) 
        & \textbf{0.0000} (48.0) 
        & 78.03 ±1.66 
        & 0.0601 (120.5) 
        & 0.2801 (179.5) 
        & \textbf{0.0000} (0.0) \\
       \midrule
        ScaleNet \((1,-1,-1)\) 
        & \textbf{0.0227} (103.0) 
        & 0.0601 (120.5) 
        & 78.59 ±1.34 
        & 0.5922 (179.5) 
        & \textbf{0.0000} (0.0) \\
       \midrule
        Dir-GNN 
        & \textbf{0.0095} (97.5) 
        & 0.2801 (179.5) 
        & 0.5922 (179.5) 
        & 78.38 ±1.30 
        & \textbf{0.0000} (0.0) \\
       \midrule
        \textbf{1}iG 
        & \textbf{0.0000} (0.0) 
        & \textbf{0.0000} (0.0) 
        & \textbf{0.0000} (0.0) 
        & \textbf{0.0000} (0.0) 
        & 71.05 ±1.91 \\
        \bottomrule
    \end{tabular}
    \caption{Wilcoxon signed-rank test results for pairwise comparisons among the top models ScaleNet\((1,1,1)\), ScaleNet\((1,-1,1)\), ScaleNet\((1,-1,-1)\), Dir-GNN, and \textbf{1}iG on the Chameleon dataset. The best model is ScaleNet with the \(\alpha, \beta, \gamma\) setting of (1,1,1).}
    \label{tab:wilcoxon_chame}
\end{table*}

%% file: table/wilcoxon/wilcoxon_squirrel.tex
            



\begin{table*}[ht]
    \centering
    \begin{tabular}{l|c|c|c|c}
        \toprule
     \textbf{Dataset: Squirrel}   & ScaleNet \((1,1,1)\) 
        & ScaleNet \((1,1,-1)\) 
        & ScaleNet \((1,-1,-1)\) 
        & Dir-GNN \\
       \midrule
        ScaleNet \((1,1,1)\) 
        & 75.31 ±1.85 
        & \textbf{0.0159} (106.0) 
        & \textbf{0.0106} (110.0)
        & 0.3085 (181.5) \\
       \midrule
        ScaleNet \((1,1,-1)\) 
        & \textbf{0.0159} (106.0) 
        & 74.75 ±1.84 
        & 0.9515 (229.0) 
        & 0.3274 (160.0) \\
       \midrule
        ScaleNet \((1,-1,-1)\) 
        & \textbf{0.0106} (110.0) 
        & 0.9515 (229.0) 
        & 74.68 ±2.09 
        & 0.3387 (185.0) \\
       \midrule
        Dir-GNN 
        & 0.3085 (181.5)
        & 0.3274 (160.0)
        & 0.3387 (185.0)
        & 75.00 ±1.91 \\
        \bottomrule
    \end{tabular}
    \caption{Wilcoxon signed-rank test results for pairwise comparisons among the top models ScaleNet\((1,1,1)\), ScaleNet\((1,1,-1)\), ScaleNet\((1,-1,-1)\), and DirGNN on the Squirrel dataset. Dir-GNN shows no significant difference from ScaleNet across the three \(\alpha, \beta, \gamma\) settings. Therefore, the best models for the Squirrel dataset are ScaleNet and Dir-GNN.}
    \label{tab:wilcoxon_squir}
\end{table*}

%% file: appendix_technical.tex
\section{Self-loops}
\subsection{Influence of Adding Self-loops}
\label{A+selfloop}
Given an adjacency matrix \( A \), the modified matrix \(\hat{A}\) with self-loops is as follows:
\[
\hat{A} = A + I,
\]
where \(I\) is the identity matrix. The products involving \(\hat{A}\) are given below:

\begin{itemize}
    \item \textbf{\(\hat{A} \hat{A}^T\)}:
        \[
    \hat{A} \hat{A}^T = (A + I)(A^T + I) = AA^T + A + A^T + I
    \]

    \item \textbf{\(\hat{A}^T \hat{A}\)}:
    \[
    \hat{A}^T \hat{A} = (A^T + I)(A + I) = A^T A + A + A^T + I
    \]

    \item \textbf{\(\hat{A} \hat{A}\)}:
    \[
    \hat{A} \hat{A} = (A + I)(A + I) = AA + 2A + I
    \]

    \item \textbf{\(\hat{A}^T \hat{A}^T\)}:
    \[
    \hat{A}^T \hat{A}^T = (A^T + I)(A^T + I) = A^T A^T + 2A^T + I
    \]
\end{itemize}

Adding self-loops to \( A \) results in a higher-scale adjacency matrix that incorporates all the corresponding directed edges of the lower-scale graphs along with additional connections from the identity matrix  \(I\).

\subsection{Generated self-loops}
\label{gen-self-loop}

\subsubsection{Proof of Generated self-loops in higher order proximity matrix}

According to DiGCN(ib) \cite{tongDigraphInceptionConvolutional2020}, the $k^{th}$-order proximity between node $i$ and node $j$ is both or either $M_{i,j}^{(k)}$ and $D_{i,j}^{(k)}$ are non-zero for:
\begin{equation*}
M^{(k)} = (\underbrace{A \cdots A}_{k-1 \text{ times}})(\underbrace{A^T \cdots A^T}_{k-1 \text{ times}})
\label{eq:Mk}
\end{equation*}
\begin{equation*}
D^{(k)} = (\underbrace{A^T \cdots A^T}_{k-1 \text{ times}})(\underbrace{A \cdots A}_{k-1 \text{ times}})
\label{eq:Dk}
\end{equation*}

\begin{proposition}
The diagonal entries would be non-zero in $M^{(k)}$ and $D^{(k)}$ (\( k \geq 2 \)) if this node has in-edge or out-edge in $A$. We call this generated self-loops.
\end{proposition}

\begin{proof}
Suppose node \( a \) has an out-edge \( a \rightarrow b \). This implies that there exists a path \( a \rightarrow b \leftarrow a \) due to the edges \( a \rightarrow b \) and \( b \leftarrow a \). As a result, the entry \( M_{a,a}^{(2)} \) in the matrix \( M^{(k)} \) will be 1, indicating a non-zero diagonal entry.

Similarly, suppose node \( a \) has an in-edge \( a \leftarrow c \). This implies that there is a path \( c \rightarrow a \) due to the edge \( a \leftarrow c \), resulting in the entry \( D_{a,a}^{(2)} \) in the matrix \( D^{(k)} \) being 1, again indicating a non-zero diagonal entry.

For all \( k \geq 2 \), \( M_{a,a}^{(k)} \) will be non-zero, as a self-loop can be traversed any number of steps and still return to the same node. The same applies to \( D_{a,a}^{(k)} \).
\end{proof}

Suppose \( M_{a,a}^{(k)} \) is non-zero, indicating that node \( a \) has a self-loop in the \( k \)-th order proximity matrix. If nodes \( d \) and \( e \) have a \( k \)-th order path meeting at node \( a \), then \( M_{d,e}^{(k+n)} \) (for \( n \geq 1 \)) will remain non-zero. This implies that higher-order matrices \( M^{(k)} \) become denser as the order increases.

If we remove the self-loops from \( M^{(k)} \), this densification effect can be eliminated. The same applies to \( D^{(k)} \).

\subsubsection{Example}
We will use an example to explain the above observation. A simple graph is shown below:

\begin{tikzpicture}
  [
    grow=down,
    sibling distance=3em,
    level distance=2em,
    edge from parent/.style={draw,latex-},
    every node/.style={font=\footnotesize},
    sloped
  ]
  \node {2}
    child { node {1} 
     child { node {6} }}
    child { node {3}
      child { node {4} }
      child { node {5} }
    }
    ;
\end{tikzpicture}

\hspace{0.1cm} 

Let's define the adjacency matrix \( A \) where non-zero entries \( A_{i,j} \) represent directed edges from node \( i \) to node \( j \). Given the directed edges (1,2), (3,2), (4,3), (5,3), (6,1), the adjacency matrix \( A \) is as follows:
\[
A=\begin{bmatrix}
  0 & 1 & 0 & 0 & 0 & 0 \\
  0 & 0 & 1 & 0 & 0 & 0 \\
  0 & 0 & 0 & 0 & 0 & 0 \\
  0 & 0 & 1 & 0 & 0 & 0 \\
  0 & 0 & 1 & 0 & 0 & 0 \\
  0 & 1 & 0 & 0 & 0 & 0
\end{bmatrix}
\]

The transpose of \( A \), denoted \( A^T \), represents the reversed edges of \( A \). Therefore, the adjacency matrix \( A^T \) is as follows:
\[
A^T=\begin{bmatrix}
  0 & 0 & 0 & 0 & 0 & 0 \\
  1 & 0 & 0 & 0 & 0 & 1 \\
  0 & 1 & 0 & 1 & 1 & 0 \\
  0 & 0 & 0 & 0 & 0 & 0 \\
  0 & 0 & 0 & 0 & 0 & 0 \\
  0 & 0 & 0 & 0 & 0 & 0
\end{bmatrix}
\]

Matrix $AA^T$, which is also ${M}^2$,  represents nodes connected with scaled edge $\rightarrow\leftarrow$ \footnote{as the backward of $\rightarrow\leftarrow$ is still $\rightarrow\leftarrow$, this scaled edge is bidirected, getting symmetric matrix.}, which are (1,3), (4,5). beacause of paths 
\begin{itemize}
    \item 1$\rightarrow2\leftarrow$3
    \item 4$\rightarrow3 \leftarrow$5
\end{itemize}

Selfloops are generated because of paths 
\begin{itemize}
    \item 1$\rightarrow2\leftarrow$1
    \item 3$\rightarrow2\leftarrow$3
    \item 4$\rightarrow3\leftarrow$4
    \item 5$\rightarrow3\leftarrow$5
\end{itemize}
Below are the matrices \( {M}^2 \) and \( \hat{M}^2 \) (\( {M}^2 \)with self-loops removed):
\[
{
\begin{array}{cc}
\overset{{M}^2}{
\begin{bmatrix}
  1 & 0 & 1 & 0 & 0 & 0 \\
  0 & 0 & 0 & 0 & 0 & 0 \\
  1 & 0 & 1 & 0 & 0 & 0 \\
  0 & 0 & 0 & 1 & 1 & 0 \\
  0 & 0 & 0 & 1 & 1 & 0 \\
  0 & 0 & 0 & 0 & 0 & 1
\end{bmatrix}}

\overset{\hat{M}^2}{
\begin{bmatrix}
  0 & 0 & 1 & 0 & 0 & 0 \\
  0 & 0 & 0 & 0 & 0 & 0 \\
  1 & 0 & 0 & 0 & 0 & 0 \\
  0 & 0 & 0 & 0 & 1 & 0 \\
  0 & 0 & 0 & 1 & 0 & 0 \\
  0 & 0 & 0 & 0 & 0 & 0
\end{bmatrix}}
\end{array}
}
\]

Matrix $(AA)(A^TA^T)$ represents undirected scaled edge $\rightarrow\rightarrow\leftarrow\leftarrow$\footnote{as its backward version is itself, it is bidirected}, which are (4,6), (5,6), because of paths:
\begin{itemize}
    \item $4\rightarrow3\rightarrow2\leftarrow2\leftarrow6$
    \item 
    $5\rightarrow3\rightarrow3\leftarrow1\leftarrow6$
\end{itemize}

One couple of $2^{nd}-order$ proximity nodes was introduced here because of self-loop in \( AA^T \), making paths:
\begin{itemize}
    \item 
    $4\rightarrow3\rightarrow2\leftarrow3\leftarrow5$
\end{itemize}

One Self-loop is generated because of path:
\begin{itemize}
    \item $6\rightarrow1\rightarrow2\leftarrow1\leftarrow6$
\end{itemize}
Below are the matrices \( {M}^3 \) and \( \hat{M}^3 \)(generated from \(\hat{M}^2 \)):

\[
\begin{aligned}
\overset{M^3}{
\begin{bmatrix}
  0 & 0 & 0 & 0 & 0 & 0 \\
  0 & 0 & 0 & 0 & 0 & 0 \\
  0 & 0 & 0 & 0 & 0 & 0 \\
  0 & 0 & 0 & 1 & 1 & 1 \\
  0 & 0 & 0 & 1 & 1 & 1 \\
  0 & 0 & 0 & 1 & 1 & 1
\end{bmatrix}} 
& 
\overset{\hat{M}^3}{ 
\begin{bmatrix}
  0 & 0 & 0 & 0 & 0 & 0 \\
  0 & 0 & 0 & 0 & 0 & 0 \\
  0 & 0 & 0 & 0 & 0 & 0 \\
  0 & 0 & 0 & 0 & 0 & 1 \\
  0 & 0 & 0 & 0 & 0 & 1 \\
  0 & 0 & 0 & 1 & 1 & 0
\end{bmatrix}}
\end{aligned}
\]

As demonstrated, both manually added self-loops to the original adjacency matrix and self-loops generated during multiplications of previous matrix  contribute to the densification of \( M^k \) (or \( D^k \)). This densification occurs because these self-loops introduce lower-order proximity edges into the higher-order proximity matrix. However, to obtain a pure \( k^{\text{th}} \)-order proximity matrix, where non-zero entries exclusively represent \( k^{\text{th}} \)-order proximities, it is crucial to remove the self-loops from each \( M^k \) and \( D^k \) before using it to compute the \( (k+1)^{\text{th}} \)-order proximity.

%% file: appendix_technical_inception.tex
\section{Digraph Inception Models}
\label{inception_Appendix}
\begin{figure*}[ht]
    \centering
 \captionsetup{skip=-20pt}
\includegraphics[width=0.8\linewidth]{fig/0_all.pdf}
\vspace{20pt} 
\includegraphics[width=\linewidth]{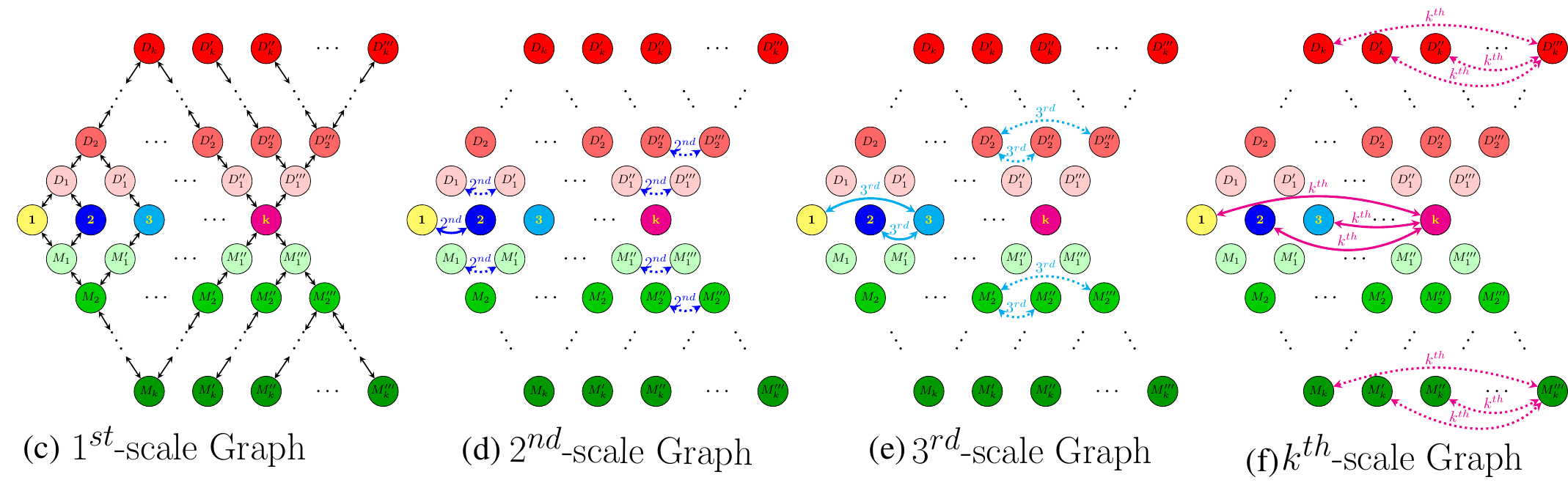}
    \caption{Edge augmentation by stacking multi-scale graphs.}
    \label{fig:scale-graph}
\end{figure*}

\subsection{Digraph Inception Networks}

 State-of-the-art Graph Neural Networks (GNNs) for homophilic graphs include Digraph Inception Networks such as DiGCN(ib)  
\cite{tongDigraphInceptionConvolutional2020} and SymDiGCN 
\cite{tongDirectedGraphConvolutional2020}, which incorporate higher-order proximity to obtain multi-scale features. 
However, these methods are based on random walks, making edge weights across various scales crucial. DiGCN(ib) uses resource-intensive eigenvalue decomposition to determine weights, while SymDiGCN relies on costly incidence normalization based on node degrees in the original graph, which limits their scalability to larger graphs.

In contrast, our method is simpler and leverages scale invariance. By transforming the graph at different scales and fusing predictions from these scaled graphs, our inception model effectively incorporates multi-scale features without the heavy computational costs incurred by existing methods.
 
Figure \ref{fig:scale-graph} illustrates the corresponding graph transformation incorporating the same scaled edges as DiGCN(ib).

DiGCN(ib) incorporates various proximities:
\begin{itemize}
    \item \textbf{$1^{st}$-order proximity} involves augmenting adverse edges, as shown in Figure \ref{fig:scale-graph}(c).
    \item \textbf{$2^{nd}$-order proximity} can be represented by the intersection or union of \( A^TA \) and \( AA^T \). The intersection includes real line edges, while the union adds dotted edges in Figure \ref{fig:scale-graph}(d). For intersection, node 1 and node 2 share 1-hop meeting node \( M_1 \) and a 1-hop diffusion node \( D_1 \), thus they are pair nodes with $2^{nd}$-order proximity. 
    \item \textbf{$3^{rd}$-order proximity} can be represented by the intersection or union of \( A^TA^TAA \) and \( AAA^TA^T \) as well. In Figure \ref{fig:scale-graph}(e), the real line edges and dotted line edges denote intersection and union, respectively.For instance, node 1 and node 3 share a 2-hop meeting node \( M_2 \) and a 2-hop diffusion node \( D_2 \). Notably, node 2 and node 3 also share these nodes. 
\end{itemize}

In the \( k \)-th scale graph, node 1 and node \( k \) have \( k \)-th order proximity as they share a \((k-1)\)-hop meeting node \( M_k \) and a \((k-1)\)-hop diffusion node \( D_k \). All nodes with lower-order proximity than \( k \) with node 1 would also have \( k \)-th order proximity with \( k \), as shown in Figure \ref{fig:scale-graph}(f), explains why higher scale graphs might get denser.

For fusion, we explore various approaches. Adding all scaled edges to the original graph results in the final augmented graph shown at Figure \ref{fig:scale-graph}(b). DiGCN(ib) uses individual GNNs for each scaled graph and simply aggregates their outputs. To fairly compare graph transformation-based inception with DiGCN(ib), we adopt their settings, with the primary difference being the assignment of scaled edge weights.

From Table 4
, \textbf{1}iG (our model) significantly outperforms DiG on the Telegram, Chameleon, and Squirrel datasets and is on par with DiG in other datasets. \textbf{1}iGi2 (our model) shows similar improvements over DiGib. DiG and \textbf{1}iG represent the $1^{st}$-scale proximity models, while DiGib and \textbf{1}iGi2 correspond to the $2^{nd}$-scale inception models.

We also tested \textbf{R}iG(i2), which assigns random weights to scaled edges within the range [0.0001, 10000].

Since the original DiG(ib) model lacks batch normalization, we experimented with both adding it and leaving it out. The results, shown in Table \ref{tab:Telegram_separated}, demonstrate that even \textbf{R}iGCN(ib) outperforms DiG(ib). This suggests that the computationally intensive procedure for assigning weights in DiGCN is less effective than simply assigning random weights, as done in RiGCN(ib), which surpasses DiGCN(ib).

We explain in Appendix \ref{DiG<RiG} why the edge weights generated by DiGCN(ib) are inferior to random weights.

Additionally, we replaced the edge weights in SymDiGCN with those in \textbf{1}ym, whose performance is comparable to SymDiGCN across all datasets.

In summary, our inception models derived from scale-invariance-based graph transformations outperform or match state-of-the-art models derived from random walks. Our methods simplify and accelerate the process by omitting edge weight calculations, yet yield better results.
\FloatBarrier

\subsection{Undesirable edge weights by DiGCN(ib)}
\begin{figure}[ht]
    \centering
    \subfigure[Accuracy with edge weights generated by DiGCN(ib): \(67.4 \pm 8.1\)]{
        \includegraphics[width=0.45\linewidth]{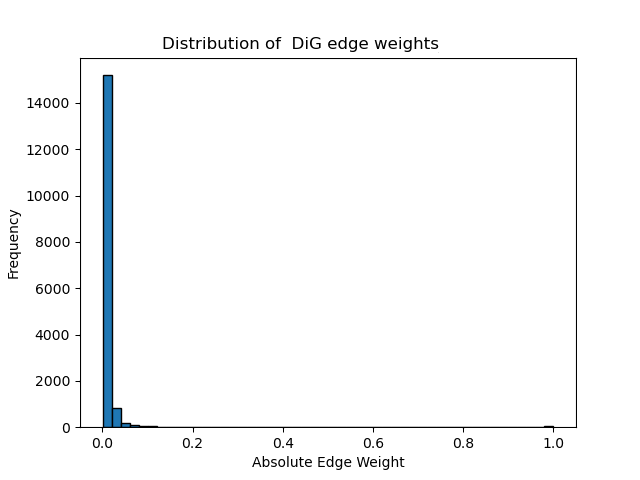}
        \label{fig:DiG_tel_weight}
    }
    \hfill
    \subfigure[Accuracy with randomly generated edge weights: \(85.2 \pm 2.5\)]{
        \includegraphics[width=0.45\linewidth]{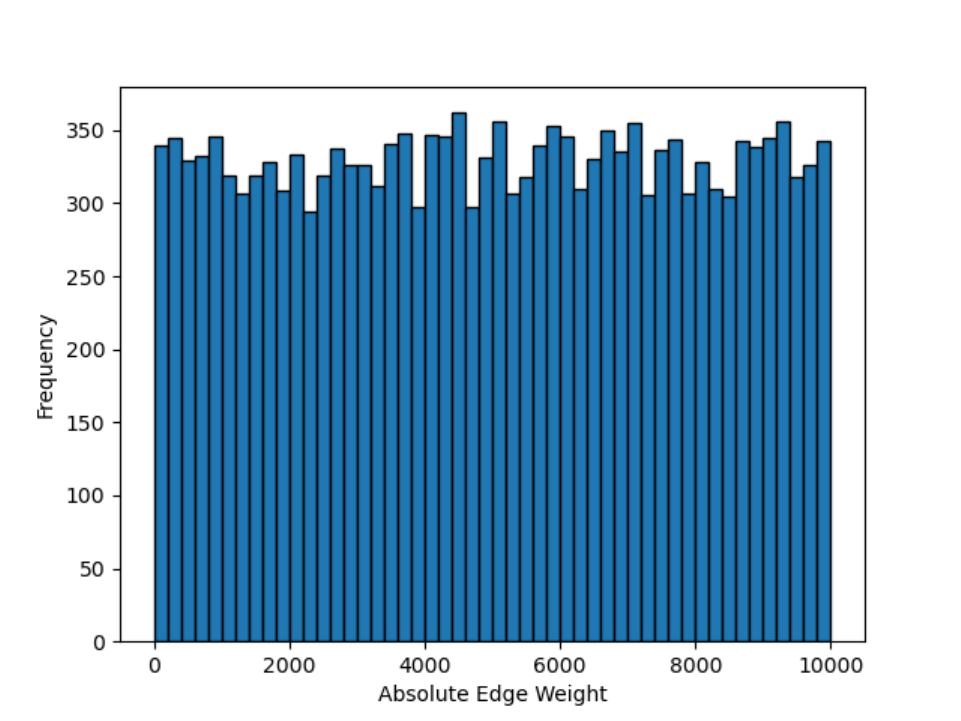}
        \label{fig:RiG_tel_weight}
    }
    \caption{DiG VS. RiG}
    \label{fig:DiG<RiG}
\end{figure}

\input{table/tele_dig_1ig_rig}
\label{DiG<RiG}
To assess the necessity of the computationally expensive edge weights used in DiGCN(ib) \cite{tongDigraphInceptionConvolutional2020}, we assigned random weights to edges, drawn from a uniform distribution in the range of \([0.0001, 10000]\). This model, named \textbf{R}iG, outperformed DiGCN(ib) on the Telegram dataset by a significant margin. The explanation for this performance is as follows:

As shown in Figure \ref{fig:DiG<RiG}, the distribution of edge weights generated by DiGCN(ib) exhibits two peaks—one around 1 and the other around 0—whereas the random weights in \textbf{R}iG follow a uniform distribution generated by a random function.

\begin{figure}[ht]
    \centering
    \subfigure[Accuracy with randomly generated edge weights (2 peaks): \(36.5 \pm 4.0\)]{
        \includegraphics[width=0.45\linewidth]{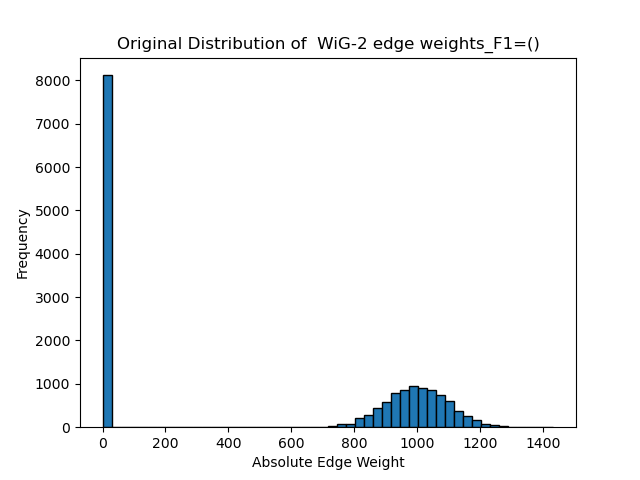}
        \label{fig:enter-label1}
    }
    \hfill
    \subfigure[Accuracy with randomly generated edge weights (3 peaks): 72.6±4.9]{
        \includegraphics[width=0.45\linewidth]{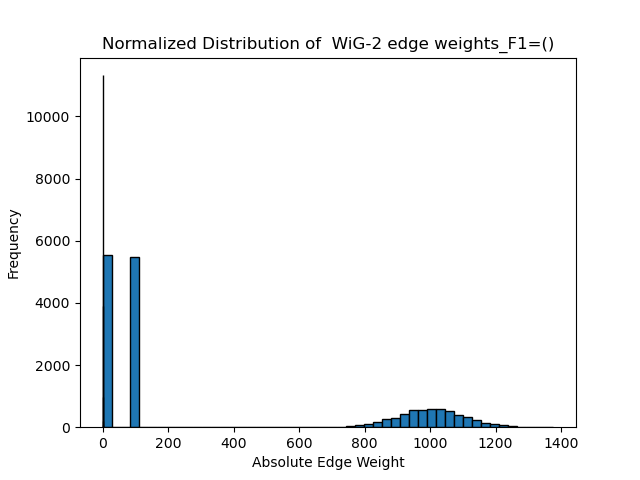}
        \label{fig:enter-label2}
    }
    \caption{Random edge weights}
    \label{fig:RandomWeight}
\end{figure}

To further investigate this, we generated sets of random weights with intentionally structured peaks, as illustrated in Figure \ref{fig:RandomWeight}. These sets included two-peak and three-peak distributions. The two-peak distribution resulted in poor performance, with an accuracy of \(36.5 \pm 4.0\), while the three-peak distribution showed significantly better performance, achieving \(72.6 \pm 4.9\), which is slightly better than DiGCN. This indicates that the specific structure of the edge weight distribution plays a crucial role in model effectiveness.

Overall, the computationally expensive edge weights generated by DiGCN(ib) are not necessarily desirable.

\FloatBarrier

%% file: table/tele_dig_1ig_rig.tex
\begin{table}[h]
    \centering
    \resizebox{0.46\textwidth}{!}{ 
    \begin{tabular}{c|c|c|c|c|c}
    \toprule
    Model & No BN & BN & Model & No BN & BN \\
    \midrule
    DiG & 67.4±8.1 & 63.0±7.6&DiGib & 68.4±6.2 & 77.4±5.1 \\
    \textbf{1}iG & 86.0±3.4 & 95.8±3.5 &\textbf{1}iGib & 86.2±3.2 & 94.2±2.7\\
    \textbf{R}iG & 85.2±2.5 & 91.0±6.3&\textbf{R}iGib & 86.4±6.2 & 86.4±6.6 \\
    \bottomrule
    \end{tabular}
    }
    \caption{Performance of Inception models on the Telegram dataset. ``BN'' indicates the addition of batch normalization to the original model. The \textbf{R}iG(ib) model assigns random weights in uniform distribution to edges within the range [0.0001, 10000], and The \textbf{1}iG(ib) model assigns weight 1 to all scaled edges.}
    \label{tab:Telegram_separated}
\end{table}

%% file: reproducibility_checklist.tex
\section{Reproducibility Checklist}
This paper:
\begin{itemize}
    \item Includes a conceptual outline and/or pseudocode description of AI methods introduced (yes/partial/no/NA)

    \textbf{yes}
    \item 
Clearly delineates statements that are opinions, hypothesis, and speculation from objective facts and results (yes/no)

\textbf{yes}
\item 
Provides well marked pedagogical references for less-familiare readers to gain background necessary to replicate the paper (yes/no)

\textbf{yes}
\end{itemize}

Does this paper make theoretical contributions? (yes/no)

\textbf{yes}
If yes, please complete the list below.

\begin{itemize}
    \item All assumptions and restrictions are stated clearly and formally. (yes/partial/no)

    \textbf{yes}
    \item 
All novel claims are stated formally (e.g., in theorem statements). (yes/partial/no)

\textbf{yes}
\item 
Proofs of all novel claims are included. (yes/partial/no)

\textbf{yes}
\item 
Proof sketches or intuitions are given for complex and/or novel results. (yes/partial/no)

\textbf{yes}
\item 
Appropriate citations to theoretical tools used are given. (yes/partial/no)

\textbf{yes}
\item 
All theoretical claims are demonstrated empirically to hold. (yes/partial/no/NA)

\textbf{yes}
\item 
All experimental code used to eliminate or disprove claims is included. (yes/no/NA)

\textbf{yes}

\end{itemize}

Does this paper rely on one or more datasets? (yes/no)

\textbf{yes}
If yes, please complete the list below.
\begin{itemize}
    \item A motivation is given for why the experiments are conducted on the selected datasets (yes/partial/no/NA)

    \textbf{yes}
    \item 
All novel datasets introduced in this paper are included in a data appendix. (yes/partial/no/NA)

\textbf{NA}
\item 
All novel datasets introduced in this paper will be made publicly available upon publication of the paper with a license that allows free usage for research purposes. (yes/partial/no/NA)

\textbf{NA}
\item 
All datasets drawn from the existing literature (potentially including authors’ own previously published work) are accompanied by appropriate citations. (yes/no/NA)

\textbf{yes}
\item 
All datasets drawn from the existing literature (potentially including authors’ own previously published work) are publicly available. (yes/partial/no/NA)

\textbf{yes}
\item 
All datasets that are not publicly available are described in detail, with explanation why publicly available alternatives are not scientifically satisficing. (yes/partial/no/NA)

\textbf{NA}
\end{itemize}

Does this paper include computational experiments? (yes/no)

\textbf{yes}
If yes, please complete the list below.

\begin{itemize}
    \item Any code required for pre-processing data is included in the appendix. (yes/partial/no).

    \textbf{yes}
    \item 
All source code required for conducting and analyzing the experiments is included in a code appendix. (yes/partial/no)

\textbf{yes}
\item 
All source code required for conducting and analyzing the experiments will be made publicly available upon publication of the paper with a license that allows free usage for research purposes. (yes/partial/no)

\textbf{yes}
\item 
All source code implementing new methods have comments detailing the implementation, with references to the paper where each step comes from (yes/partial/no)

\textbf{yes}
\item 
If an algorithm depends on randomness, then the method used for setting seeds is described in a way sufficient to allow replication of results. (yes/partial/no/NA)

\textbf{yes}
\item 
This paper specifies the computing infrastructure used for running experiments (hardware and software), including GPU/CPU models; amount of memory; operating system; names and versions of relevant software libraries and frameworks. (yes/partial/no)

\textbf{yes}
\item 
This paper formally describes evaluation metrics used and explains the motivation for choosing these metrics. (yes/partial/no)

\textbf{yes}
\item 
This paper states the number of algorithm runs used to compute each reported result. (yes/no)

\textbf{yes}
Analysis of experiments goes beyond single-dimensional summaries of performance (e.g., average; median) to include measures of variation, confidence, or other distributional information. (yes/no)

\textbf{yes}
\item 
The significance of any improvement or decrease in performance is judged using appropriate statistical tests (e.g., Wilcoxon signed-rank). (yes/partial/no)

\textbf{yes}

\item 
This paper lists all final (hyper-)parameters used for each model/algorithm in the paper’s experiments. (yes/partial/no/NA)

\textbf{yes}
\item 
This paper states the number and range of values tried per (hyper-) parameter during development of the paper, along with the criterion used for selecting the final parameter setting. (yes/partial/no/NA)

\textbf{yes}
\end{itemize}